%% file: MAIN.tex
\documentclass[sigconf]{acmart}

\usepackage{booktabs} 
\usepackage{todonotes}
\usepackage{tabularx}
\usepackage{multirow} 
\usepackage{color}
\usepackage{makecell}

\newcommand{\q}[1]{\lq\lq{}{}#1\rq\rq{}{}}

\copyrightyear{2019} 
\acmYear{2019} 
\setcopyright{acmcopyright}
\acmConference[SAC '19]{The 34th ACM/SIGAPP Symposium on Applied Computing}{April 8--12, 2019}{Limassol, Cyprus}
\acmBooktitle{The 34th ACM/SIGAPP Symposium on Applied Computing (SAC '19), April 8--12, 2019, Limassol, Cyprus}

\begin{document}

\title[Distant Supervision for Predicting and Understanding Entity Linking Difficulty]{Same but Different: Distant Supervision for Predicting and Understanding Entity Linking Difficulty}

\author{Renato Stoffalette Jo\~{a}o}
\orcid{0000-0003-4929-4524}
\affiliation{
 \institution{L3S Research Center, Leibniz University of Hannover}
  \streetaddress{Appelstra{\ss}e 9A}
  \city{Hannover}
  \state{Germany}
}
\email{joao@L3S.de}

\author{Pavlos Fafalios}
\orcid{0000-0003-2788-526X}
\affiliation{
 \institution{L3S Research Center, Leibniz University of Hannover}
  \streetaddress{Appelstra{\ss}e 9A}
  \city{Hannover}
  \state{Germany}
}
\email{fafalios@L3S.de}

\author{Stefan Dietze}
\affiliation{%
 \institution{GESIS - Leibniz Institute for the Social Sciences}
  \city{K\"oln}
  \state{Germany}
}
\additionalaffiliation{
 \institution{L3S Research Center, Leibniz University of Hannover}
  \streetaddress{Appelstra{\ss}e 9A}
  \city{Hannover}
  \state{Germany}
}
\email{stefan.dietze@GESIS.org}

\begin{abstract}
Entity Linking (EL) is the task of automatically identifying entity mentions in a piece of text and resolving them to a corresponding entity in a reference knowledge base like Wikipedia. 
There is a large number of EL tools available for different types of documents and domains, yet EL remains a challenging task where the lack of precision on particularly ambiguous mentions often spoils the usefulness of automated disambiguation results in real applications. 
A priori approximations of the difficulty to link a particular entity mention can facilitate flagging of critical cases as part of semi-automated EL systems, while detecting latent factors that affect the EL performance, like corpus-specific features, can provide insights on how to improve a system based on the special characteristics of the underlying corpus. 
In this paper, we first introduce a consensus-based method to generate difficulty labels for entity mentions on arbitrary corpora. 
The difficulty labels are then exploited as training data for a supervised classification task able to predict the EL difficulty of entity mentions using a variety of features. 
Experiments over a corpus of news articles show that EL difficulty can be estimated with high accuracy, revealing also latent features that affect EL performance. Finally, evaluation results demonstrate the effectiveness of the proposed method to inform semi-automated EL pipelines. 
\end{abstract}

\keywords{Entity Linking; Named Entity Recognition and Disambiguation; Supervised Classification; Distant Supervision}

\maketitle

\input{introduction}

\input{motivation}

\input{methodology}

\input{experiments}

\section{Conclusions and Future Work}
\label{sec:conclusion}

We have investigated the novel problem of detecting and understanding EL difficulty. 
To this end, we first introduced a method to generate difficulty labels for entity mentions in arbitrary corpora, by utilising agreement and disagreement sets obtained through state-of-the-art EL systems. As shown in the evaluation results, our approach to detect difficult to link mentions as part of a semi-automated EL pipeline can considerably improve the performance of state-of-the-art EL tools, by enabling the efficient prediction of critical cases which require manual labelling.
For example, the accuracy of a popular EL system (Ambiverse) is increased by 6 percentage points when 10\% of the recognised mentions, labelled as \textit{HARD} by our method, are manually judged. 

Subsequently, we introduced a set of features which can be used within a distantly supervised model for predicting difficult to link mentions on the fly, for cases where no labels can be assigned by the proposed labelling method or when real time analysis is needed. Evaluation results on the NYT corpus showed that difficulty labels can be predicted with high precision ($>$0.83) and recall ($>$0.72) even using limited amounts of training data, while recall can be further improved using a balanced training dataset. Our multifeature model highly outperforms baselines using the mention length or the number of mention's candidate entities only, demonstrating that context-specific features as well as temporal features are required in order to achieve reasonable performance.
In addition, this prediction task can be used for detecting latent characteristics that affect EL performance on a given corpus. 
In the NYT corpus, for example, we saw that the position of the mention in the document characterises many HARD cases because long-tail entities (article authors) are usually listed at the last position.

Future work is concerned with reflecting more complex features, such as \textit{lexical diversity} \cite{dillard2002persuasion} or \textit{document fluency} \cite{horne2016expertise}. We also plan to investigate the effectiveness of common oversampling methods (like SMOTE \cite{chawla2002smote}) as well as cost-sensitive classifiers and more balancing techniques, focusing on further increasing the prediction performance for the minority class (\textit{HARD}).

\vspace{-1mm}
\begin{acks}
This work was partially supported by CNPq (Brazilian National Council for Scientific and Technological Development) under grant GDE No. 203268/2014-8 and the European Commission for the ERC Advanced Grant ALEXANDRIA under grant No. 339233.
\end{acks}

\bibliographystyle{ACM-Reference-Format}
\bibliography{bibliography}

\end{document}

%% file: introduction.tex
\section{Introduction}
Entity linking (EL), or named entity recognition and disambiguation (NERD), is the task of determining the identity of entity mentions in texts, thereby linking a mention to an entity within a reference Knowledge Base (KB), such as Wikipedia. EL is a crucial task of relevance for a wide variety of applications, such as Web search, information retrieval, or document classification. Usually, high precision (P) and recall (R) is required if EL results are to have a positive impact on any such application.

However, EL remains a challenging task. Systems differ along multiple dimensions and are evaluated over different datasets \cite{shen2015entity}, while their performance differs significantly across domains and corpora \cite{roder2017gerbil}. EL difficulty varies per corpus but also with each individual mention, where previous work has shown that mentions which are difficult to link often share common characteristics \cite{hoffart2012kore}. Typical examples include highly ambiguous mentions where a large number of potential candidates exists, mentions of long-tail entities which are not well represented in KBs, such as local public figures, or mentions whose meaning changes over time. 

Given that automated EL pipelines never reach perfect P/R on arbitrary corpora, human judgements are often required to improve automatically generated EL results \cite{cheng2015summarizing,singh2016discovering,demartini2012zencrowd}. Therefore, estimating a priori the difficulty of linking a particular mention can facilitate high P/R systems, e.g. by flagging critical mentions which require manual judgements as part of semi-automated EL approaches. Such approaches utilise the scalability of automated linkers wherever possible and benefit from the precision of human judgements to handle challenging cases. 
In this context, in particular the widely used practice of applying state-of-the-art EL systems out of the box calls for methods that enable detecting difficult to link mentions as well as latent characteristics that affect the EL performance, thus addressing the strong context-specific nature of EL.

In this paper, we first introduce an automated method to generate difficulty labels (\textit{HARD}, \textit{MEDIUM}, \textit{EASY}) for entity mentions in an arbitrary corpus. The proposed method utilises agreement and disagreement measures obtained by applying state-of-the-art EL systems on the given corpus. Experimental results demonstrate the effectiveness of this labelling strategy on improving the performance of semi-automated EL, by enabling the efficient prediction of critical cases which require manual labelling (e.g., from domain experts or through crowdsourcing).

To detect characteristics that determine the difficulty of a mention to be linked correctly, as well as to allow predicting EL difficulty on-the-fly (e.g., for cases where real-time analysis is needed, or when no labels can be assigned using the proposed labelling method), we exploit the generated difficulty labels as training data for a multi-class classification task able to predict the EL difficulty of entity mentions using a diverse feature set. 
Through an extensive feature analysis we investigate the importance of different types of features, inspired by previous work as well as by the observed characteristics of difficult-to-link mentions. 

We apply our approach to the New York Times (NYT) corpus \cite{sandhaus2008new}
and find that the position of the mention in the document, the size of the sentence containing the mention, and the frequency of the mention in the document (all related to the mention's context) are the three most useful features for predicting EL difficulty in our experiments, while temporal features also contribute.
In addition, we demonstrate that EL difficulty can be estimated on the fly with high precision ($>$0.83) and recall ($>$0.72) even using a limited amount of the available training data (below 25\% of the original data), while recall can be further improved using a balanced training dataset.
While to the best of our knowledge no works exist which address this prediction task, we compare our configurations to two baselines which utilise few but highly predictive features (number of candidate entities, mention length) and show superior performance of our multifeature approach. In a nutshell, we make the following contributions: 

\begin{itemize}
    \item We introduce an automated approach to generate difficulty labels which relies on agreement information among different EL systems. The generated labels can be used to improve semi-automated EL, as direct indicators or through distant supervision. 
    \item We propose a novel approach, feature sets and classifiers for predicting EL difficulty as well as for detecting latent, corpus-specific characteristics that affect EL performance. 
\end{itemize}

The rest of the paper is organised as follows:
Section \ref{sec:motiv} motivates the problem and discusses related works. 
Section \ref{sec:assignment} introduces the proposed method to assign difficulty labels. 
Section \ref{sec:classification} describes the features used in our multi-class classification task. 
Section \ref{sec:casestudy} reports experimental results on predicting and understanding EL difficulty. 
Section \ref{sec:impact} shows how the proposed method can improve semi-automated EL. 
Finally, Section \ref{sec:conclusion} concludes the paper and discusses interesting directions for future research.

%% file: motivation.tex
\section{Motivation and Related Work}
\label{sec:motiv}

\subsection{Motivation}
\label{subsec:motivation}
Whereas both users and applications of automatically generated entity annotations usually require high performance, in particular, high precision, EL remains a challenging task, where no single system has yet emerged as de-facto-standard.
Evaluations using the GERBIL benchmark \cite{roder2017gerbil}, a framework that compares EL systems over a large number of ground truth datasets, have shown that their performance is highly affected by the characteristics of the datasets, like the number of entities per document, the document length, the total number of entities, or the salient entity types \cite{usbeck2015evaluating}. This demonstrates that, the widely used practice of applying state-of-the-art EL systems out of the box, i.e. without corpus-specific training, usually does not provide the best performance.  

In particular, wrongly linked mentions often share certain common characteristics, where typical examples include: 
i) highly ambiguous mentions which often have a large number of candidate entities and/or are short (e.g. family names \q{Brown} or \q{Williams});  
ii) mentions of long-tail entities (often not represented in reference KBs, e.g. regional politicians);  
iii) mentions of entities where the respective meaning evolves significantly over time (e.g. \q{Germany} before or after 1990, or \q{President of the US});
iv) mentions of entities where the popularity, and hence prior probability, of disambiguation candidates changes significantly over time (like \q{Amazon} in 1980 or 2018); v) mentions which are prone to partial matching, such as location names (e.g. \q{Madrid} which may refer to the city or the football club Real Madrid depending on the context).

These features underline the corpus-specific nature of EL difficulty. 
For these reasons, when applying any state-of-the-art system to an arbitrary corpus, estimating the actual quality of the produced annotations remains challenging. In addition, independent of the overall performance, real-world applications which utilise annotations call for quality standards which cannot necessarily be met by automated EL approaches alone.
Thus, estimating a priori the difficulty of linking a particular mention can facilitate high precision systems, e.g. by flagging critical mentions which require manual judgements as part of semi-automated EL approaches \cite{cheng2015summarizing,singh2016discovering,demartini2012zencrowd}.

\vspace{-2mm}
\subsection{Related Work}
\label{subsec:rw}

\input{related_work}

%% file: related_work.tex
The survey in \cite{shen2015entity} presents a thorough overview and analysis of the main approaches to EL. 
Below we report works based on the document or mention features they consider which are of relevance for both EL as well as the problem addressed in our work.

Although the majority of systems can operate over any type of document, there are systems designed for specific \textit{document types} like tweets \cite{guo2013link,fang2014entity}, queries ~\cite{blanco2015fast}, and web lists ~\cite{shen2012liege}. Similarly, there are systems tailored for specific \textit{domains}, like medicine \cite{rajani2017stacking} or scientific publications in general ~\cite{jana2017wikim}. 
The \textit{document context} is exploited by most of the state-of-the-art EL systems for collectively disambiguating the mentions in a given text \cite{ferragina2010tagme,hoffart2011robust,moro2014entity,shen2015entity}. The idea is to consider the coherence among the candidate entities for all the detected mentions.
The \textit{time} and \textit{location} aspects have been considered in \cite{fang2014entity} for improving the EL performance on microblogs. 

The \textit{length} of a mention and its \textit{number of candidate entities} are considered two of the main characteristics that affect EL difficulty and have been taken into consideration for creating ground truth datasets of difficult to annotate entity mentions. For example, the KORE50 dataset contains entity mentions having a large number of candidate entities, while the WP dataset replaces full names with family names only, thereby reducing the mention's length \cite{hoffart2012kore}. 
EL difficulty is also affected by the presence of \textit{long-tail entities}, i.e. entities that do not exist in the reference KB or that are not well represented. Regarding the former, \cite{zheng2010learning,ratinov2011local} leverage supervised machine learning techniques to predict unlinkable entity mentions.
Regarding entities that are not well represented in KBs (like emerging entities), \cite{cheng2015summarizing,singh2016discovering} propose  methods that incorporate human feedback during the EL process.

Finally, \cite{brasoveanu2018framing} describes a taxonomy to frame common errors in gold standards as well as errors generated by EL systems. The taxonomy was tested in an experimental environment where errors were manually identified and classified by human annotators.

To our knowledge, our work is the first that aims at computing difficulty labels for EL in arbitrary corpora, as well as predicting EL difficulty and detecting corpus-specific characteristics that can influence the performance of state of the art EL systems.

%% file: methodology.tex
\section{Consensus-based Labelling}
\label{sec:assignment}

\subsection{Problem Formulation}
Let $D$ be a corpus of documents, e.g., a set of news articles, covering the time period $T_D$. Consider also a contemporary KB $K$, for instance Wikipedia, describing information for a set of entities $E$. 
The output of applying EL on the documents of $D$ is a set of annotations of the form $\langle d, m, p, e\rangle$, where $d$ is a document in $D$, $m$ is an entity mention in $d$ (a word or a sequence of words), $p$ is the position of $m$ in $d$, and $e$ is an entity in $E$ that determines the identity of $m$. 

We now define the problem of determining the difficulty in linking a mention $m$ to an entity in $K$ as a multi-class classification problem where $m$ is assigned to one of the following classes: 
\vspace{-0.5mm}
\begin{itemize}
    \item \textit{HARD}: Difficult to disambiguate mention (state of the art EL systems usually fail to find the correct link)
    \item \textit{EASY}: Easy to disambiguate mentions (state of the art EL systems almost always find the correct link)
    \item \textit{MEDIUM}: All other cases (neither EASY nor HARD)
\end{itemize}
\vspace{-1mm}

Below we describe an automated approximation strategy to assign these difficulty labels on entity mentions of an arbitrary corpus. 

\vspace{-1mm}
\subsection{Labelling Process}
\label{subsec:labelling}

We propose to use freely available state-of-the-art EL systems $\langle s_1, .. , s_n\rangle$ which operate on the same reference KB $K$ (e.g., Wikipedia 2016) and are applied to the same corpus $D$. The degree of agreement of all systems $s_i$ is then used as indicator of the EL difficulty. 

In particular, assuming $n=3$ systems, three sets of entity links are produced ($A_1$, $A_2$, and $A_3$).
To generate the labels we consider only the commonly recognised entities, i.e., the mentions for which all three systems provide a link, which may or may not be the same. The set $A$ of common entity annotations has elements of the form $\langle d, m, p, e_1, e_2, e_3\rangle$ where $d$ is the document, $m$ is the mention, $p$ is the position of $m$ in $d$, and $e_1$, $e_2$, and $e_3$ are the entities provided by $s_1$, $s_2$, and $s_3$, respectively.

A mention $m_i$ is assigned with the \textit{HARD} label if all three systems disagree, i.e. each one provides a link to a different entity $e_j$. The intuition is that in this case, at least 2/3 systems failed to find the correct entity. Formally, for $n=3$ the set of HARD annotations $A^H$ is defined as: 
\begin{equation}
  A^H = \{\langle d, m, p, e_1, e_2, e_3\rangle \in A ~|~ e_1 \neq e_2 \neq e_3\}  
\end{equation}

As \textit{EASY} we consider the cases where all systems agree on the same mention, i.e., all provide the same entity link. Formally: 
\begin{equation}
A^{E} = \{\langle d, m, p, e_1, e_2, e_3\rangle \in A ~|~ e_1 = e_2 = e_3\}
\end{equation}

As \textit{MEDIUM} we consider all other cases:
\begin{equation}
A^{M} = \{a \in A ~|~ a \notin A^H \wedge a \notin A^E\}
\end{equation}
i.e., cases where exactly 2/3 systems provide the same entity.

It is obvious that the above labelling process can provide wrong approximations since it assumes that if the systems provide the same entity link then this link is correct. Our assumption is that, in particular the EASY class might contain false positives to a certain degree, e.g. when all systems agree on the same but wrong entity. 
In Section \ref{subsec:evalsetup:labeling} we provide evaluation results of the quality of class assignments obtained through our approach, suggesting a precision of more than 93\% on average given our experimental setup.

An additional limitation arises from the fact that this labelling method requires mentions to be recognised by all the considered systems, i.e., it cannot provide labels for mentions recognised by only one or two of the systems. As shown in our experiments (Sect. \ref{subsec:evalsetup:labeling}), the common mentions are less than 30\% of the total mentions recognised by each system, thus we need to predict the EL difficulty of all other mentions. Furthermore, the  efficiency of this labelling method depends on the efficiency of the used systems, thus it might not be applicable for cases where real-time analysis is needed or large amounts of documents are to be annotated. 

\section{Learning Entity Linking Difficulty}
\label{sec:classification}

To address the issues of the aforementioned labelling strategy, supervised classification can be used to predict EL difficulty. In particular, a distantly supervised classification model may be trained using the proposed labelling strategy in order to learn to predict the linking difficulty of arbitrary entity mentions.
For this, we need a diverse set of features which covers different aspects of EL difficulty. 

Inspired by previous works as well as by the observed characteristics of difficult to link mentions which are not correctly disambiguated through state-of-the-art systems (cf. Section \ref{subsec:motivation}), we consider features of the following three categories: 
i) \textit{mention-based} (features of the mention itself),
ii) \textit{document-based} (features of the document containing the mention), and
iii) \textit{temporal} (features that consider the temporal evolution of either the mention or the document containing the mention).
Below we detail each of them, while a summary is given in Table \ref{tbl:features}.

\begin{table}[!h]
\vspace{-2mm}
\centering
\renewcommand{\arraystretch}{0.85}
\setlength{\tabcolsep}{3.5pt}
\caption{Summary of the considered features.}
\vspace{-3mm}
\label{tbl:features}
\footnotesize
\begin{tabular}{>{\raggedright\arraybackslash}p{0.9cm} >{\raggedright\arraybackslash}p{0.9cm} p{5.5cm}}	   \toprule
   Category & Notation &  Description  \\
   \midrule
		\multirow{1}{0pt}{Mention}	 		 &  $m_{len}$   & Num of mention's characters (length). \\ 
		     		     &  $m_{words}$   & Num of mention's words. \\ 
		     		     &  $m_{freq}$   & Num of mention's occurrences in the doc (frequency).   \\ 
		     		     &  $m_{df}$  & Num of docs in the corpus containing at least one occurrence of the mention.     	 \\ 
	    			     &  $m_{cand}$  & Num of mention's candidate entities in a reference KB. \\ 
	    			     &  $m_{pos}$  & Mention's normalised position in the doc (num of chars from the start of the doc / total num of doc's chars). \\ 
	    			     &  $m_{sent}$  & Num of chars of the sentence containing the mention. \\ 
	    \midrule
					
    	\multirow{1}{0pt}{Document}         &  $d_{words}$    & Num of words in the document containing the mention. \\
			 		     &  $d_{topic}$  & Main topic discussed in the document containing the mention (e.g., SPORTS, or POLITICS). \\
			 		     &  $d_{ents}$    & Num of entity mentions recognised in the document containing the mention. \\ 		
		\midrule
			 		  
    	Temporal 		 &  $t_{age}$    & The distance (age) of the doc's publication date from the date of the reference KB. 	        \\ 	
    	                 &  $t_{df}$   & Number of docs containing at least one occurrence of the mention, published  within $k$ intervals from the doc's publication date (e.g., $+/-$ 6 months). \\
	    	 		     &  $t_{j_{min}}$ / $t_{j_{max}}$ / $t_{j_{avg}}$    & Min, max and avg Jaccard similarity of the mention's top-K similar words (computed using Word2Vec) for all pairs of consecutive time periods of fixed granularity. \\ 
	    \bottomrule
\end{tabular}
\label{tab:features}
\vspace{-3.5mm}
\end{table}

\vspace{-1mm}
\subsection{Mention-based Features}

\textbf{Mention length ($m_{len}$):} the number of mention's characters. Short mentions are usually more ambiguous compared to long mentions (e.g., {\em Adams} vs {\em Schwarzenegger}). 

\noindent
\textbf{Mention words ($m_{words}$):} the mention's number of words. Unigram mentions are usually more ambiguous than mentions with more than one word (e.g., {\em John} vs {\em John McCain}). 

\noindent
\textbf{Mention frequency ($m_{freq}$):} the number of mention occurrences within the document. More occurrences imply that the document is closely related to the mention, thus the context of the mention is more likely to be related to the actual mention.  

\noindent
\textbf{Mention document frequency ($m_{df}$):} the number of documents in the corpus $D$ containing at least one occurrence of the mention. Higher $m_{df}$ implies popularity of the term(s), suggesting that more context is available about this mention. 

\noindent
\textbf{Mention candidate entities ($m_{cand}$):} the number of candidate entities in the reference KB. The articles in Wikipedia (the most common reference KB) contain hyperlinks with anchor texts pointing to entities, making it an important source for mining mention and entity relations. For a mention $m$ we select as candidate entities those that appear as link destinations for $m$. A higher number of candidate entities indicates a more ambiguous mention. 

\noindent
\textbf{Mention's normalised position ($m_{pos}$):} the mention's normalised position in the document, computed as the number of characters from the start of the document divided by the total number of document's characters. 
Entities that appear early in the document are usually salient and representative for the document, indicating more representative context to facilitate their disambiguation.

\noindent
\textbf{Mention's sentence size ($m_{sent}$):} The number of characters of the sentence containing the mention, specifically the length of the text between two punctuation marks containing the mention (considering only the punctuation marks ".", "!", "?", ";").
An EL system may exploit the sentence containing the mention for disambiguating the entity, where larger sentences indicate more representative context for a particular mention. 

\vspace{-1mm}
\subsection{Document-based Features}

\noindent
\textbf{Document size ($d_{words}$):} the number of words of the document containing the mention. Small documents do not provide much context information what hinders precise disambiguation of its entity mentions.

\noindent
\textbf{Document topic ($d_{topic}$):} the main topic (subject) discussed in the document containing the mention, selected from a predefined list of topics (like SPORTS, POLITICS, etc.). This information can be obtained either through an automated document classification algorithm or directly through the document's metadata (if such information is available). The difficulty to disambiguate mentions varies among topics, for instance, related to the specificity of the topic or the prevalence of long-tail entities.

\noindent
\textbf{Document's recognised entities ($d_{ents}$):} the total number of entities recognised in the document containing the mention. State of the art EL systems jointly disambiguate the entities in a document, e.g. by considering the linking structure in a reference KB. Thus, more recognised entities provide more contextual information enabling more precise disambiguation. 

\vspace{-1mm}
\subsection{Temporal Features}

\textbf{Document publication age ($t_{age}$):} the distance of the document's publication date from the date of the reference KB (measured based on a fixed time interval, e.g., years). For example, if Wikipedia 2016 is the reference KB, a document of 2000 has age 16 while a document of 1990 has age 26. Mentions in old documents are more difficult to disambiguate since temporally distant entities are less well-represented or their context may have changed (e.g., linking the mention {\em Ronaldo} in a today's article vs in an article of 1990's).

\noindent
\textbf{Mention's temporal document frequency ($t_{df}$):} the number of documents containing the mention, published  within $k$ intervals from the publication date of the document (e.g., $+/-$ 6 months). Higher $t_{df}$ means that the corresponding entity was popular during that particular time period, indicating the context of the mention is more likely to refer to the respective mention.

\noindent
\textbf{Mention's semantics stability ($t_{j_{min}}$, $t_{j_{max}}$, $t_{j_{avg}}$):} the minimum, maximum, and average Jaccard similarity coefficient of the mention's top-K similar words for all pairs of consecutive time intervals. 
The documents are grouped into a sequence of $n$ time interval-specific subsets based on a fixed time granularity $\Delta$ (e.g., year) and a Word2Vec Skipgram model \cite{mikolov2013efficient} is trained for each group of documents (resulting in $n$ different models). Given a mention, we retrieve its top-K similar words in each interval using the Word2Vec models and compute the Jaccard similarity of these sets of words for all pairs of consecutive time periods. We consider the minimum, maximum and average Jaccard similarity among all pairs.
These three features consider the semantic evolution of terms, where the meaning of a term may change over time or the prior probability of a mention-entity link significantly changes due to temporal events 
(e.g., {\em Germany} is likely to refer to Germany's national football team during international football tournaments).

%% file: experiments.tex
\section{Experimental evaluation}
\label{sec:casestudy} 

We evaluate the performance of supervised classification models on learning EL difficulty in a given corpus. The models make use of the proposed labelling strategy (cf. Section \ref{sec:assignment}) and feature set (cf. Section \ref{sec:classification}) for i) predicting the EL difficulty of entity mentions, and ii) detecting corpus characteristics that affect the EL performance.

\vspace{-1mm}
\subsection{Setup}

\subsubsection{Corpus}

We used the New York Times (NYT) Annotated Corpus \cite{sandhaus2008new} which contains over 1.8 million articles published by the NYT between 1987 and 2007, covering a wide range of topics (like sports, politics, arts,  business) and diverse content formats (like long texts, short notices, corrections, and headlines). The number of articles per year ranges from 79,077 (in 2007) to 106,104 (in 1987).

\vspace{-1.5mm}
\subsubsection{Labelling}
\label{subsec:evalsetup:labeling}

We implemented the proposed labelling strategy (cf. Section \ref{subsec:labelling}) using the EL systems \textit{Ambiverse} (previously AIDA) \cite{hoffart2011robust}, \textit{Babelfy} \cite{moro2014entity}, and \textit{TagMe} \cite{ferragina2010tagme}. 
In all three systems, we used Wikipedia 2016 as the common reference KB. 
For Ambiverse, we used its public Web API with the default configuration. 
For Babelfy, we used a local deployment and a configuration suggested by the Babelfy developers\footnote{The configuration is available at: \url{https://goo.gl/NHXVVQ}}. For TagMe we used a local deployment with the default configuration and a confidence threshold of 0.2 to filter out low quality annotations.
We examined the performance of each system on the widely-used CoNLL-TestB ground truth \cite{hoffart2011robust}. Ambiverse achieved 81\% precision and 65\% recall, Babelfy 81\% precision and 68\% recall, and TagMe 79\% precision and 53\% recall. The performance of the systems is very close to the one reported in the literature for the same dataset.

The number of commonly recognised mentions among the three systems is 11,876,437, which corresponds to 30\%, 11\% and 21\% of the total mentions recognised by Ambiverse, Babelfy and TagMe, respectively. 
We see that our labelling strategy cannot assign labels to a large number of mentions which have not been recognised by all three  systems, thus we need to predict the linking difficulty of these mentions. 
From the common mentions, 340,238 (2.9\%) are HARD (all systems disagree with each other), 9,070,517 (78.6\%) are EASY (all systems provide the same entity) and 2,465,682 (21.4\%) are MEDIUM (2/3 systems provide the same entity). We notice that the labels are highly unbalanced: the number of HARD cases is much smaller than the number of EASY and MEDIUM cases. 

{\em Quality of generated labels.}
First, we examined if the HARD mentions are indeed hard for all three systems or if there is one showing consistently high performance on these cases.   
We manually produced the ground truth for a random sample of 500 HARD cases. Ambiverse, Babelfy and TagMe managed to find the correct entity in 24\%, 16\% and 31\% of the cases, respectively. 
We notice that the joint effectiveness of all systems is low, supporting our labelling strategy.  
Then we examined the precision of the EASY and MEDIUM labels. 
We randomly selected 200 mentions from the EASY class and for each one we manually examined if the entity provided by the three systems is correct. The accuracy for this subset is 95\%, i.e. only 5\% of the mentions have been wrongly classified as EASY. Regarding the MEDIUM class, we randomly selected 200 mentions and tested if the two systems that agree provide the correct entity (if not, then these mentions can be considered HARD). In this case we found that 12\% of the mentions have been wrongly classified as MEDIUM. Considering that the majority (78.6\%) of the not-HARD cases are EASY (following the original unbalanced distribution), we expect an error rate of MEDIUM and EASY labels of less than 7\%.

The generated annotations as well as the ground truths of the aforementioned qualitative evaluation are made publicly available.\footnote{\url{http://l3s.de/~joao/SAC2019/}}

\vspace{-1.5mm}
\subsubsection{Balancing \& Sampling}
To cater for the highly uneven class distribution, we experimented with both \textit{unbalanced} and \textit{balanced} training data.
The unbalanced training dataset maintains the actual class distribution as observed in the data, while the balanced training dataset randomly undersamples the majority classes (all classes have the same number of training instances). 

In order to compare the impact of dataset size, we examined different stratified sampling approaches: 
i) \textit{SAMPLE25} (random 25\% stratified sample of the full dataset), 
ii) \textit{SAMPLE10} (random 10\% stratified sample of the full dataset), and
iii) \textit{SAMPLE1} (random 1\% stratified sample of the full dataset).
In all the experiments we applied 10-fold cross validation, using 90\% of the instances for training and the remaining 10\% for testing. Note that in the balanced datasets, undersampling of the training data of the majority classes is part of the cross validation, i.e. the test data is always unbalanced. 

\vspace{-1.5mm}
\subsubsection{Classification Models}
Considering the scale of the data as well as the features, we apply the following classifiers:
i) \textit{Naive Bayes} (a classifier that assumes that the likelihood of the features follows a Gaussian distribution),
ii) \textit{Logistic Regression} (a classifier that models the label probability based on a set of independent variables), 
iii) \textit{Decision Tree} (a classifier that successively divides the features space to maximise a metric), and
iv) \textit{Random Forest} (a classifier that utilises an ensemble of uncorrelated decision trees).

\vspace{-1.5mm}
\subsubsection{Baselines and multifeature approach}
While some related works deal with the prediction of unlinkable mentions \cite{shen2015entity}, no state-of-the-art baselines do exist which address the classification task proposed in our work. 
We follow the assumption that the ambiguity of a mention is strongly dependent on the available candidates in a KB as well as the mention length. These two features are known to strongly influence EL difficulty and have been used for creating gold standards of difficult test cases \cite{hoffart2012kore}.
Thus, we consider the following baselines:
i) \textsc{CandidNum} (classification using only the feature $m_{cand}$), and
ii) \textsc{MentLength} (classification using only the feature $m_{len}$).
We compare the performance of these baselines with a \textsc{MultiFeature} classifier which considers all the features described in Section \ref{sec:classification} (cf. Table \ref{tbl:features}).

\vspace{-1.5mm}
\subsubsection{Configurations}
Depending on the corpus (NYT in our case), some of the features need to be configured accordingly. For the \textit{document topic} ($d_{topic}$), we exploited the taxonomic classification provided by NYT. Each document was assigned to one of the following topics: \textit{Arts, Automobiles, Books, Business, Education, Health, Home and Garden, Job Market, Magazine, Movies, New York and Region, Obituaries, Real Estate, Science, Sports, Style, Technology, Theatre, Travel, Week in Review, World, Miscellaneous}. For the \textit{document publication age} ($t_{age})$, we used \textit{year} as the time interval. For the \textit{mention's temporal document frequency} ($t_{df}$), we used $k$=6 months as the interval. For the \textit{mention's semantics stability} ($t_j$), we used K=50 and $\Delta$=year, while in the Word2Vec Skipgram model we set the default setting as also used in \cite{mikolov2013distributed} (300 dimensions, 5 words window size). 
Regarding the examined classifiers, we used their default configuration in WEKA \cite{hall2009weka}. 

\vspace{-1.5mm}
\subsubsection{Evaluation Metrics}
To evaluate the performance of the different classifiers, we consider \textit{Precision} (P) (the fraction of the correctly classified instances among the instances assigned to the class), \textit{Recall} (R) (the fraction of the correctly classified instances among all instances of the class), and \textit{F1 score} (the harmonic mean of P and R).
We report the prediction performance per class as well as the macro average performance, to ensure that the size of each class has no impact on the representativeness of our metrics.

\subsection{Results}
\label{sec:evalresults}

\subsubsection{Classification Performance}

Table \ref{tbl:overall} summarises the overall results of the baselines (\textsc{Can\-did\-Num}, \textsc{Me\-nt\-Length}) and our multifeature approach (\textsc{Mu\-lti\-Fea\-tu\-re}) for the SAMPLE25 dataset. The table shows the macro averages of our performance metrics for both the unbalanced and balanced training dataset. 

In all cases, we observe that using the proposed \textsc{MultiFeature} approach with a Random Forest classifier provides the best results, outperforming the baselines.
Paired t-tests with $\alpha$-level 5\% indicate that this improvement is statistically significant in all cases.
With respect to the baselines, we observe that \textsc{CandidNum} (number of mention's candidate entities) outperforms \textsc{MentLength} (mention's length). 
We also note that the unbalanced dataset achieves higher macro average F1 score compared to the balanced dataset (0.76 vs 0.60). In more detail, using the unbalanced training dataset we obtain higher macro average precision compared to the balanced dataset (0.83 vs 0.58), however recall is lower (0.72 vs 0.76). 

Tables \ref{tbl:detailed} shows the detailed performance per class for both the \textit{unbalanced} and \textit{balanced} training datasets. Looking at the \textsc{MultiFeature} results of Random Forest  for the unbalanced dataset, we notice that, as expected, the majority class EASY achieves high scores (0.92 precision and 0.97 recall). The MEDIUM class also performs very well (0.83 precision and 0.71 recall), while the HARD class achieves high precision (0.75) but lower recall (0.46). Regarding the HARD class, we see that using the balanced dataset recall is highly increased to 0.84, but precision drops to 0.21. 
We also observe that, when using \textsc{MentLength} with the unbalanced dataset, all classifiers learn to assign all instances to the majority class.  

\begin{table}[]
\vspace{-2mm}
\centering
\caption{Overall prediction performance (macro average) using SAMPLE25.}
\vspace{-3mm}
\renewcommand{\arraystretch}{0.65}
\setlength{\tabcolsep}{3.5pt}
\small
\label{tbl:overall}
\begin{tabular}{@{}llllllll@{}}
\toprule
\multirow{2}{*}{\textsc{\textbf{Method}}} & 
\multirow{2}{*}{\textsc{\textbf{Model}}} & \multicolumn{3}{c}{\textsc{\textbf{Unbalanced}}} & \multicolumn{3}{c}{\textsc{\textbf{Balanced}}} \\ 
 \cmidrule(l){3-5} \cmidrule(l){6-8} & & P & R & F1 & P & R & F1 \\ \midrule
\multirow{4}{*}{\textsc{CandidNum}} 
                & \textsc{Naive Bayes} & 0.38 & 0.35 & 0.34 & 0.37 & 0.40 & 0.32  \\
                & \textsc{Logistic Regr.} & 0.31 & 0.33 & 0.30 & 0.43 & 0.41 & 0.35  \\
                & \textsc{Decision Tree}  & 0.74 & 0.47 & 0.50 & 0.48 & 0.61 & 0.47  \\
                & \textsc{Random Forest} & 0.74 & 0.47 & 0.50 & 0.48 & 0.61 & 0.47  \\
 \midrule
\multirow{4}{*}{\textsc{MentLength}} 
                & \textsc{Naive Bayes} & 0.25 & 0.33 & 0.29 & 0.36 & 0.42 & 0.26  \\
                & \textsc{Logistic Regr.} & 0.25 & 0.33 & 0.29 & 0.37 & 0.44 & 0.31  \\
                & \textsc{Decision Tree}  & 0.25 & 0.33 & 0.29 & 0.42 & 0.47 & 0.40  \\
                & \textsc{Random Forest} & 0.25 & 0.33 & 0.29 & 0.42 & 0.47 & 0.39 \\
 \midrule
\multirow{4}{*}{\textsc{MultiFeature}} 
                & \textsc{Naive Bayes} & 0.42 & 0.41 & 0.41 & 0.43 & 0.49 & 0.41  \\
                & \textsc{Logistic Regr.} & 0.45 & 0.36 & 0.35 & 0.43 & 0.50 & 0.40  \\
                & \textsc{Decision Tree}  & 0.74 & 0.69 & 0.71 & 0.56 & 0.74 & 0.59 \\
                & \textsc{Random Forest} & \textbf{0.83} & \textbf{0.72} & \textbf{0.76} & \textbf{0.58} & \textbf{0.76} & \textbf{0.60}  \\
 \bottomrule
\end{tabular}
\vspace{-3mm}
\end{table}

\begin{table*}[]
\centering										\vspace{-2mm}				
\caption{Prediction performance per class using SAMPLE25.}
\vspace{-3mm}						
\renewcommand{\arraystretch}{0.65}
\setlength{\tabcolsep}{3.5pt}
\small
\label{tbl:detailed}
\begin{tabular}{@{}llcccccccccccccccccc@{}}
\toprule
\multirow{3}{*}{\textsc{\textbf{Method}}} & \multirow{3}{*}{\textsc{\textbf{Model}}} & \multicolumn{9}{c}{\textsc{\textbf{Unbalanced Training}}}  & \multicolumn{9}{c}{\textsc{\textbf{Balanced Training}}} \\ \cmidrule(l){3-11} \cmidrule(l){12-20} 
    \multicolumn{1}{c}{} & \multicolumn{1}{c}{}   & \multicolumn{3}{c}{\textsc{\textbf{Hard}}} & \multicolumn{3}{c}{\textsc{\textbf{Medium}}} & \multicolumn{3}{c}{\textsc{\textbf{Easy}}} & \multicolumn{3}{c}{\textsc{\textbf{Hard}}} & \multicolumn{3}{c}{\textsc{\textbf{Medium}}} & \multicolumn{3}{c}{\textsc{\textbf{Easy}}} \\ \cmidrule(l){3-5} \cmidrule(l){6-8} \cmidrule(l){9-11}    \cmidrule(l){12-14}  \cmidrule(l){15-17} \cmidrule(l){18-20}

\multicolumn{1}{c}{} & \multicolumn{1}{c}{}  & P & R & F1 & P & R & F1 & P & R & F1 & P & R & F1 & P & R & F1 & P & R & F1  \\ \midrule
\multirow{5}{*}{\textsc{CandidNum}} 																	
 & \textsc{Naive Bayes} 	& 0	& 0	& 0	& 0.37	& 0.11	& 0.17 & 0.78	& 0.96	& 0.86 & 0.06 & 0.32 & 0.10 & 0.26 & 0.03 & 0.05 & 0.80 & 0.86 & 0.83 \\
 & \textsc{Logistic Regr.} 	& - & 0	& - & 0.16 & 0.02 & 0.04 & 0.76 & 0.97 & 0.85  & 0.05 & 0.33 & 0.09 & 0.41 & 0.09 & 0.14 & 0.82 & 0.83 & 0.82 	\\
 & \textsc{Decision Tree}  	& 0.67 & 0.08 & 0.14 & 0.73 & 0.35 & 0.47 & 0.83 & 0.98 & 0.90  & 0.10 & 0.65 & 0.17 & 0.43 & 0.48 & 0.45 & 0.92 & 0.69 & 0.79 	\\
 & \textsc{Random Forest} 	& 0.67 & 0.08 & 0.14 & 0.72 & 0.35 & 0.47 & 0.83 & 0.97 & 0.90  & 0.09 & 0.66 & 0.16 & 0.44 & 0.47 & 0.45 & 0.92 & 0.69 & 0.79 	\\
                        \midrule																	
                        
\multirow{5}{*}{\textsc{MentLength}}																	
& \textsc{Naive Bayes} & - & 0 & - & - & 0 & - & 0.76 &	1 &	0.87 &  0.05 & 0.72 & 0.08 & 0.14 & 0.12 & 0.13 & 0.88 & 0.43 & 0.58 \\
& \textsc{Logistic Regr.} & - &	0 & - & - & 0 & - & 0.76 & 1 & 0.87 & 0.05 & 0.61 & 0.09 & 0.18 & 0.13 & 0.15 & 0.87 & 0.57 & 0.69 \\
& \textsc{Decision Tree}  & - &	0 & - & - & 0 & - & 0.76 & 1 & 0.87 & 0.06 & 0.40 & 0.10 & 0.33 & 0.38 & 0.35 & 0.87 & 0.64 & 0.74 \\
& \textsc{Random Forest}  & - &	0 & - & - & 0 & - & 0.76 & 1 & 0.87 &  0.06 & 0.41 & 0.10 & 0.33 & 0.38 & 0.35 & 0.87 & 0.63 & 0.73 \\
   \midrule				                        
                        
\multirow{5}{*}{\textsc{\textsc{MultiFeature}}}    																
& \textsc{Naive Bayes}   & 0.04 & 0.01 & 0.02 & 0.40 & 0.39 & 0.39 & 0.82 & 0.84 & 0.83  & 0.06 & 0.47 & 0.11 & 0.37 & 0.31 & 0.34 & 0.86 & 0.70 & 0.77 	\\
& \textsc{Logistic Regr.}   & 0	& 0 & 0 & 0.58 & 0.11 & 0.18 & 0.78 & \textbf{0.98} & 0.87  & 0.07 & 0.49	& 0.12 & 0.33 & 0.39 & 0.36 & 0.88 & 0.63 & 0.74	\\
& \textsc{Decision Tree}    & 0.55 & 0.43 & 0.48 & 0.76 & 0.69 & 0.73 & \textbf{0.92} & 0.95 & 0.94  & 0.20 & 0.79 & 0.32 & 0.55 & 0.62 & 0.58 & \textbf{0.95} & 0.81 & 0.87 	\\
& \textsc{Random Forest}   & \textbf{0.75} & \textbf{0.46} & \textbf{0.57} & \textbf{0.83} & \textbf{0.71} & \textbf{0.77} & \textbf{0.92} & 0.97 & \textbf{0.95}  & \textbf{0.21} & \textbf{0.84} & \textbf{0.34} & \textbf{0.57} & \textbf{0.63} & \textbf{0.60} & \textbf{0.95} & \textbf{0.82} &\textbf{ 0.88} 	\\
    \bottomrule																
\end{tabular}	
\vspace{-3mm}
\end{table*}

\vspace{-1mm}
\subsubsection{Influence of Dataset Size}
\begin{figure}[!h]
\vspace{-2mm}
\centering
\fbox{\includegraphics[width=2.5in]{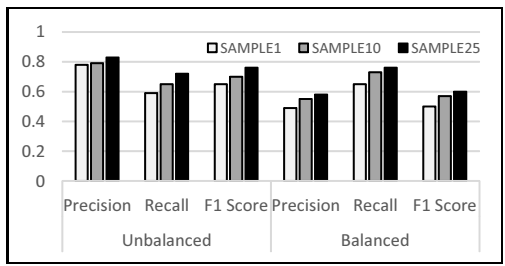}}
\vspace{-2.5mm}
\caption{Influence of dataset size on prediction performance (macro average) using Random Forest.}
\label{fig:sizeEffectMULTI}
\vspace{-3mm}
\end{figure}

Figure \ref{fig:sizeEffectMULTI} shows the performance of our multifeature Random Forest classifier for different size of training data. As expected, the use of more training instances results in better performance. For instance, the F1 score using the unbalanced dataset increases from 0.65 (1\% sample) to 0.7 (10\% sample) and 0.76 (25\% sample). 
We also notice that the dataset size affects recall more than precision. 
In general, even when using only 1\% of the dataset, precision is quite high using the unbalanced dataset ($0.78$).

\vspace{-1mm}
\subsubsection{Feature Analysis}
To evaluate the usefulness of individual features, we compute the Mean Decrease Impurity (MDI) per feature, applied to the Random Forest model (the best performing classifier). MDI quantifies the importance of a feature by measuring how much each feature decreases the impurity in a tree, where in our analysis we considered information gain (entropy). We computed MDI using both the unbalanced and balanced SAMPLE25 datasets.

\begin{figure}[!h]
\vspace{-2mm}
\centering
\fbox{\includegraphics[width=2.6in]{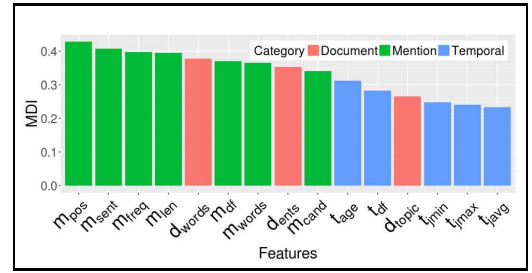}}
\vspace{-3mm}
\caption{Attribute importance (Mean Decrease Impurity) per feature for SAMPLE25.}
\label{fig:mdi}
\vspace{-3mm}
\end{figure}

Figure \ref{fig:mdi} shows the average MDI score per feature (differences between the unbalanced and balanced datasets were minor). Surprisingly, the most useful feature is the mention's normalised position ($m_{pos}$), followed by the size of the sentence containing the mention ($m_{sent}$), the frequency of the mention in the document ($m_{freq}$), and the mention length ($m_{len}$). We see that 3/4 of these features are related to the mention context.
By inspecting several articles of the corpus we notice that a particular cause for this observation is the fact that author names are commonly added at the end of an article ($m_{pos} \approx 1$). These entity mentions usually appear only once in the article ($m_{freq} = 1$) and usually correspond to long-tail entities (with no Wikipedia entry). Hence such mentions tend to be of the HARD class. In addition, entities that appear early in the document (small $m_{pos}$ value) are usually representative for the document, indicating more representative context which in turn facilitates their disambiguation. 
With regard to the high MDI score of $m_{sent}$ (size of the sentence containing the mention), we noticed that several articles with HARD cases provide long lists of long-tail entities (like the roster of a local team, or congress representatives). In such cases, the size of the sentence containing the mention is usually very small. 

In general, we notice that the most important features are the mention-based features whereas temporal features impact the performance to a lesser extent (having though an MDI score of $>0.2$). With respect to the document-based features, the document size ($d_{words}$) is the most useful (5th among all features), validating our hypothesis that small documents do not provide much context information and this hinders precise disambiguation of its mentions. With regard to temporal features, the publication age of the document containing the mention ($t_{age}$) has the largest MDI value, while the three features related to the mention's semantics stability ($t_{j_{min}}$, $t_{j_{max}}$, $t_{j_{avg}}$) have the lowest contribution. 

\begin{figure}
\centering
\fbox{\includegraphics[width=2.7in]{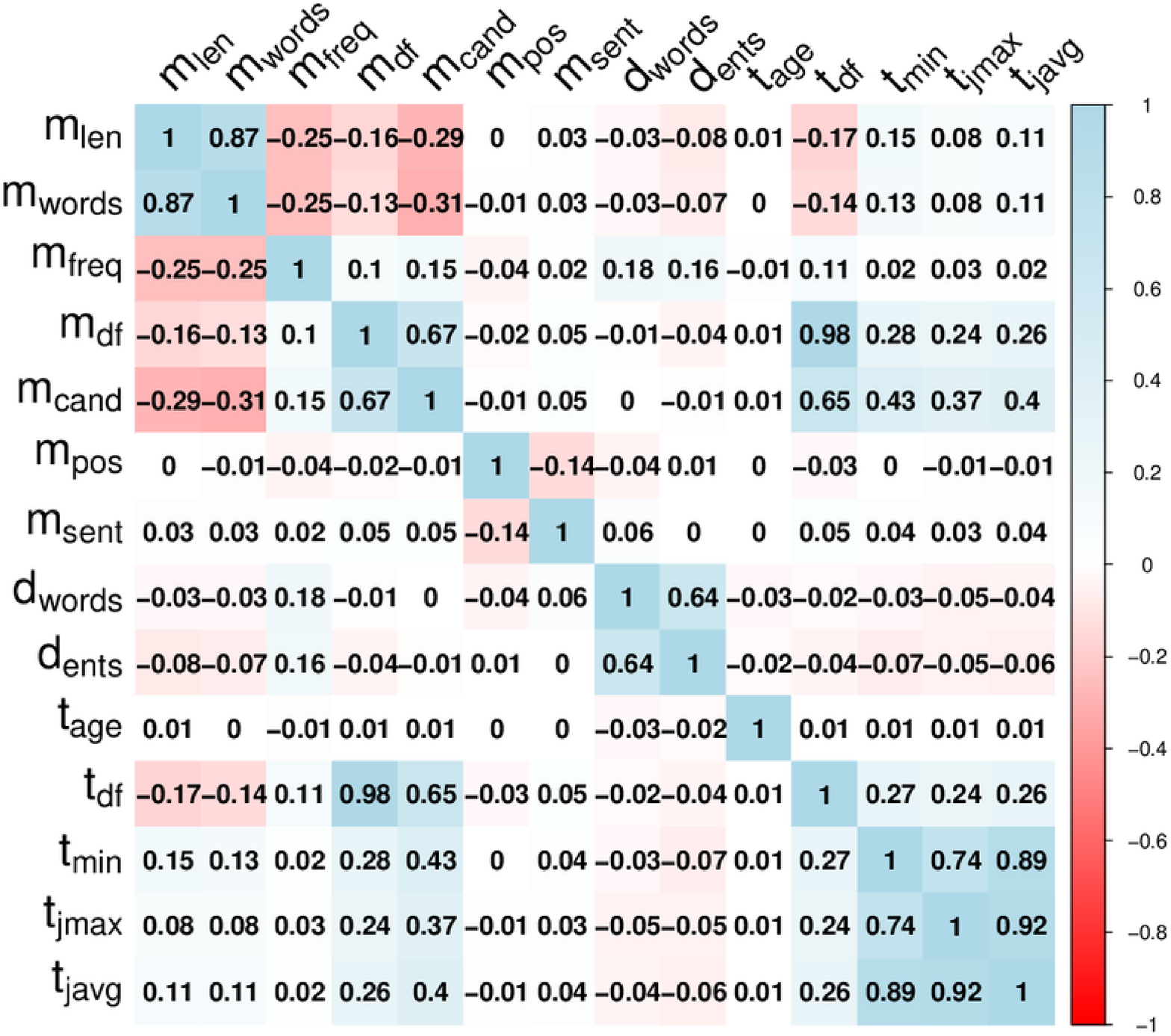}}
\vspace{-3mm}
\caption{Correlation among features (Pearson's r).}
\label{fig:corr}
\vspace{-4mm}
\end{figure}

Note that a low MDI value indicates that, either the feature is not important or it is highly correlated with one or more of the other features. To assess correlation among the features, we examined the correlation matrix  using Pearson's correlation coefficient. The results are depicted in Figure \ref{fig:corr} (we do not consider the nominal feature $d_{topic}$). 

The comparatively high correlation between the mention's document frequency ($m_{df}$) and temporal document frequency ($t_{df}$) indicates that one of both likely is redundant, thus we can consider only $t_{df}$ to avoid parsing the entire corpus. The high correlation among the min, max and average mention's semantics stability ($t_{j_{min}}, t_{j_{max}}, t_{j_{avg}}$) suggests that, in the case of our corpus, we may consider only one of these features. 
As expected, the number of mention's characters ($m_{len}$) is strongly correlated with the number of mention's words ($m_{words}$) (more words means longer strings), and the document size ($d_{words}$) has a strong correlation with the number of document's recognised entities  ($d_{ents}$) (large documents usually imply more recognised entities). 
An interesting correlation is that of the mention's number of candidate entities ($m_{cand}$) with the mention's document frequency ($m_{df}$) and temporal document frequency ($t_{df}$). A possible explanation is the following: small values of $m_{df}$ (or $t_{df}$) may imply a less popular term which might correlate with a smaller amount of disambiguation candidates ($m_{cand}$). This correlation may also explain the surprisingly low MDI value of $m_{cand}$ (as shown in Figure \ref{fig:mdi}).  

We examined the performance of Random Forest without considering the features $t_{df}, t_{j_{max}}, t_{j_{avg}}, m_{words}, d_{ents}$ (which are highly correlated to other features). Using SAMPLE25 and the unbalanced training dataset, we obtain the following macro average performance: P = $0.83$, R = $0.71$, F1 = $0.76$. We observe that the results are almost the same with the ones reported for the entire feature set. Using the smaller SAMPLE1 dataset, we obtain P = $0.77$, R = $0.58$, F1 = $0.64$. Again the performance is similar to the all-features approach (slightly worse). These results illustrate that we can omit some features that are expensive to compute and which have a strong correlation with other, less expensive features. 

\section{Impact on Entity Linking}
\label{sec:impact}
To demonstrate the application of detecting difficult to link mentions, we assess the overall performance of semi-automated EL pipelines, where human annotators are guided by our classification task to complement system-generated entity links with manual annotations in particularly challenging cases. 
We used three state-of-the-art EL systems (\textit{Ambiverse}, \textit{Babelfy}, and 
\textit{TagMe}), configured as described in the previous section (cf. Section \ref{subsec:evalsetup:labeling}) and using Wikipedia 2016 as the common reference KB. 
We consider a corpus for which gold standard annotations are provided, in particular the CoNLL-TestB ground truth \cite{hoffart2011robust}, and applied the proposed method to generate difficulty labels.

From the commonly recognised mentions among the systems that also exist in the ground truth (2,471 mentions), 
we select a random set of \textit{N} {\tt DIFFICULT} mentions (labelled as \textit{HARD} by our method) and consider that a human provides the correct link for these mentions. We do the same for a random set of \textit{N} mentions predicted as \textit{HARD} by a Random Forest classifier ({\tt PRED.DIFFICULT}).\footnote{We trained the classifier using the full unbalanced training dataset of CoNLL and all features described in Section \ref{sec:classification} apart from the three temporal features and the document topic (CoNLL does not provide this information).}
In both cases, if the number of \textit{HARD} mentions is smaller that \textit{N}, we fill up with random \textit{MEDIUM} mentions.
We compute the accuracy of the three systems (number of correctly linked mentions / total number of mentions) in both cases and compare the results with the accuracy of the systems on the same dataset before the human intervention ({\tt BEFORE}), as well as with two baselines: i) one which randomly selects mentions for manual judgement ({\tt RANDOM}), and ii) one which selects mentions based on their number of candidate entities, starting with the mentions having the more candidate entities ({\tt CANDIDATES}).   
In all cases, for selecting the mentions to manually judge, we run the experiment 10 times for 10 different random sets of selected mentions, and we report the average results.

\begin{figure}[!h]
\centering
\includegraphics[width=3.36in]{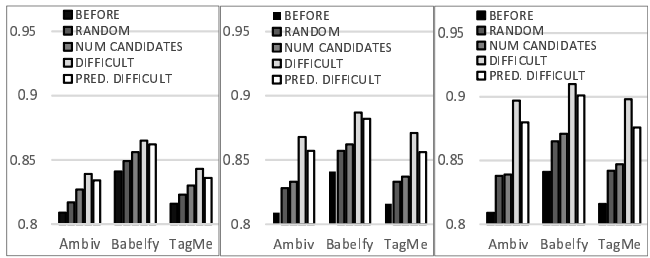}
\vspace{-6mm}
\caption{Effect of human feedback on the accuracy of semi-automated EL systems for different proportion of human judgements: 5\% (left), 10\% (middle), and 15\% (right).}
\label{fig:simulation}
\vspace{-2mm}
\end{figure}

Figure \ref{fig:simulation} depicts the results for different proportion of manually judged entity links: 5\% of the mentions (\textit{N} = 124) (left), 10\% of the mentions (\textit{N} = 247) (middle), and 15\% of the mentions (\textit{N} = 371) (right).
We notice that the proposed method ({\tt DIFFICULT}) highly improves the performance of all systems, while the improvement is considerably higher compared to the two baselines. 
Ambiverse, for instance, improves its accuracy from 0.81 to 0.84, 0.87, and 0.9, using 5\%, 10\%, and 15\%, respectively, of the mentions for manual judgement. Moreover, using a pre-trained classifier ({\tt PRED.DIFFICULT}), the improvement is again high and very close to the {\texttt DIFFICULT} case (outperforming again the two baselines). For example, Ambiverse improves its accuracy from 0.81 to 0.83, 0.86, and 0.88, using 5\%, 10\%, and 15\%, respectively, of the mentions for manual judgement.
These results demonstrate the effectiveness of our strategy on selecting difficult to link mentions (possible disambiguation errors).

